\pdfoutput=1

\documentclass[11pt]{article}

\usepackage[]{EMNLP2023}

\usepackage{times}
\usepackage{latexsym}

\usepackage[T2A,T1]{fontenc}
\usepackage[utf8]{inputenc}
\usepackage[russian,english]{babel}

\usepackage{microtype}

\usepackage{inconsolata}

\title{Open Language Data Initiative: Advancing Low-Resource Machine Translation for Karakalpak}

\author{Mukhammadsaid Mamasaidov \\
  Tahrirchi \\
  \texttt{m.mamasaidov@tahrirchi.uz} \\\And
  Abror Shopulatov \\
  Tahrirchi \\
  \texttt{a.shopulatov@tahrirchi.uz} \\}

\begin{document}
\maketitle
\begin{abstract}
This study presents several contributions for the Karakalpak language: a FLORES+ devtest dataset translated to Karakalpak, parallel corpora for Uzbek-Karakalpak, Russian-Karakalpak and English-Karakalpak of 100,000 pairs each and open-sourced fine-tuned neural models for translation across these languages. Our experiments compare different model variants and training approaches, demonstrating improvements over existing baselines. This work, conducted as part of the Open Language Data Initiative (OLDI) shared task, aims to advance machine translation capabilities for Karakalpak and contribute to expanding linguistic diversity in NLP technologies.
\end{abstract}

\section{Introduction}

The Karakalpak language, a member of the Turkic language family, is primarily spoken in the Republic of Karakalpakstan, an autonomous region within Uzbekistan, Central Asia. Current estimates suggest a native speaker population to be around 900,000 individuals \cite{ethnologue_kaa}. Linguistically, Karakalpak is an agglutinative language which belongs to the Kipchak branch of the Turkic language family and shares close affinities with Kazakh and Nogai \cite{berdimuratovetaldauletov1979}.

As a low-resource language, Karakalpak presents significant challenges in the field of natural language processing, particularly in machine translation. Major translation platforms such as Google Translate \cite{GoogleTranslate} currently do not offer support for this language as of the time of writing this paper, underscoring the need for dedicated research and development in this area. 

This study, conducted as part of the Open Language Data Initiative (OLDI) shared task, presents fine-tuned neural models for Karakalpak translation, a fine-tuned version the No Language Left Behind (NLLB) model \cite{nllb-22}. In line with OLDI's goals of expanding language resources, we release several key contributions:

\begin{enumerate}
    \item A FLORES+ devtest dataset \cite{nllb-22} translated to Karakalpak
    \item Parallel corpora for Uzbek-Karakalpak, Russian-Karakalpak and English-Karakalpak of 100,000 pairs each \footnote{\url{https://huggingface.co/datasets/tahrirchi/dilmash}}
    \item Open-sourced fine-tuned neural models for translation across Uzbek, Russian, English and Karakalpak languages \footnote{\url{https://huggingface.co/collections/tahrirchi}}
    \item Scripts for Latin-Cyrillic transliteration for Karakalpak
\end{enumerate}

Our research aims to advance the state of machine translation for Karakalpak, contributing to the broader OLDI objective of improving natural language processing capabilities for low-resource languages. This work demonstrates how shared tasks like OLDI can drive progress in expanding linguistic diversity in NLP technologies.

\section{Related work}

\begin{table*}[t]
\centering
\small
\begin{tabular}{|p{0.48\textwidth}|p{0.48\textwidth}|}
\hline
\textbf{English} & \textbf{Karakalpak} \\
\hline
According to Japan's nuclear agency, radioactive caesium and iodine has been identified at the plant. & Yaponiya yadro agentligi maǵlıwmatlarına kóre, stanciyada radioaktiv ceziy hám yod bar ekenligi anıqlanǵan. \\
\hline
The result of plotting analysis will be posted to a public website. & Syujet analiziniń nátiyjesi ǵalabalıq veb-saytqa jaylastırıladı. \\
\hline
The station's web site describes the show as "old school radio theater with a new and outrageous geeky spin!" & Stanciya veb-saytında show "jańa hám ádettegiden basqasha ájáyıp aylandıratuǵın eski mektep radio teatrı!" dep táriyiplenedi. \\
\hline
\end{tabular}
\caption{Examples from the FLORES+ dataset for English-Karakalpak language pair}
\label{tab:flores-examples}
\end{table*}

The field of machine translation for low-resource languages has experienced significant advancement with the advent of the No Language Left Behind (NLLB) model families. These innovative models demonstrate the capability to facilitate translation across more than 200 languages, leveraging extensive collections of online corpora.

Another notable multilingual translation model is MADLAD-400 \cite{kudugunta2024madlad}, which extends the capabilities of large language models to cover 400 languages, including many low-resource languages and Karakalpak. This model represents a significant step forward in expanding the reach of machine translation to a broader range of linguistic communities.

In the specific context of Karakalpak machine translation, several notable efforts have been made. A prominent example is the Apertium platform \cite{forcada2011apertium}, a rule-based machine translation system designed for low-resource languages. Utilizing finite-state algebra and rule-based methodologies, Apertium has developed morphological analyzers and spell-checking tools for Karakalpak\footnote{\url{https://github.com/apertium/apertium-uzb-kaa}}. Furthermore, it has produced machine translation systems for language pairs for Uzbek-Karakalpak, Kazakh-Karakalpak, and Tatar-Karakalpak.

A recent contribution to the Karakalpak translation comes from the Turkic Interlingua (TIL) team \cite{mirzakhalov2021turkic}, who introduced a model specifically trained on Turkic languages and corpora, with Karakalpak included in its linguistic scope. This initiative not only enhances the translation capabilities for Karakalpak but also contributes to the broader landscape of Turkic language processing. Additionally, the team has made significant strides in corpus development, introducing parallel corpora for numerous Turkic language pairs, including those involving Karakalpak.

Moreover, a proprietary online translation service for Karakalpak exists at \url{https://from-to.uz/}. To provide a comprehensive evaluation of Karakalpak machine translation capabilities, we will assess this tool's performance using its API, comparing it with our proposed models. This comparison will offer insights into both open-source and commercial solutions for low-resource language translation.

To our best knowledge, these developments collectively represent important steps towards improving machine translation capabilities for Karakalpak and other low-resource languages within the Turkic language family.

As an additional benchmark, we will include Claude-3.5-sonnet, a commercial large language model (LLM) with multilingual capabilities. While not specifically designed for Karakalpak translation, Claude-3.5-sonnet represents the current state of general-purpose language models and can provide valuable insights into how well such models perform on low-resource language tasks.

\section{Datasets}

\subsection{FLORES+ Devtest Dataset}

This study introduces the Karakalpak FLORES+ devtest dataset, which comprises 1012 sentences translated from English to Karakalpak. The FLORES+ datasets, derived from Wikimedia content, have been widely employed in the evaluation of foundational models within the NLLB family.

This dataset was developed under the auspices of the Open Language Data Initiative (OLDI). Two annotators were responsible for the translation of a devtest split, with subsequent cross-verification to ensure accuracy. The Karakalpak translations adhere to the most recent iteration of the Latin script orthography (see Table \ref{tab:flores-examples}).

The Karakalpak orthography has experienced multiple changes recently. Both Latin and Cyrillic scripts are utilized, with the Latin script, introduced in 1995, undergoing several revisions. Notable modifications occurred in 2009 and 2016, with the latter replacing digraphs with diacritic letters to overcome previous limitations. Conversion scripts for Cyrillic and older Latin versions to the current system are available on GitHub\footnote{\url{https://github.com/tahrirchi/kaa-scripts}}.

\subsection{Training data}

The training dataset comprises diverse parallel corpora sourced from multiple domains, including news articles, literary works, lexicographic resources, and educational materials. Specifically, the corpus encompasses on average across three languages:

\begin{itemize}
\item 23\% sentences from news sources
\item 34\% sentences from books (novels, non-fiction)
\item 24\% sentences from bilingual dictionaries
\item 19\% sentences from school textbooks
\item 4,000 English-Karakalpak pairs from Gatitos Project \cite{jones2023gatitos}
\end{itemize}

In total, the dataset consists of 100,000 sentence pairs for Uzbek-Karakalpak, Russian-Karakalpak, and English-Karakalpak each, making 300,000 pairs in total. Since there were too few bitexts with English, we decided to create English-Karakalpak dataset by translating Russian sentences from the Russian-Karakalpak dataset to English using Claude 3.5 Sonnet (See Appendix \ref{sec:claude-prompt}). To promote further research and development in this field, we have made these corpora publicly available. 

\subsection{Data Mining Process}

For mining parallel sentences, we apply only local mining, when we are sure that parallel sentences are to be mined from the translations of the same book, document or article. For alignment, we use LaBSE embeddings, although Karakalpak is not a supported language in LaBSE. We found that due to similarities of Karakalpak to already included Uzbek and Kazakh languages, LaBSE performed well for aligning sentences so we skipped this step.

The sentence alignment method we use is identical to the one applied for Erzya, as described by \cite{dale-2022-first}. We utilize LaBSE (Language-agnostic BERT Sentence Embedding) \cite{feng2020language} to generate embeddings for each sentence pair. To calculate the alignment score, we first determine the cosine similarity between these embeddings. We then adjust this similarity by multiplying it with a length ratio - specifically, the length of the shorter sentence divided by the length of the longer sentence.

Using dynamic programming, we identify the sequence of sentence pairs that maximizes the total similarity score. Finally, we apply a threshold to filter out low-scoring alignments.

\section{Translation Experiments}

\subsection{Model Training}

For our experiments, we utilized the nllb-200-distilled-600M model, which is a transformer-based neural machine translation model with an encoder-decoder architecture. This model comprises 12 layers and employs the following approach: the source and target languages are indicated by the first tokens of the encoder and decoder inputs, respectively. This architecture allows the model to process and translate between numerous language pairs. The training process for our experiments consisted of several key steps:

\subsubsection{Tokenizer Preparation}
Initially, we trained a SentencePiece \cite{kudo2018sentencepiece} tokenizer on an expanded monocorpus of approximately 300,000 Karakalpak sentences with a total of 16,000 vocabulary length. We decided to train a separate tokenizer because we hypothesized that the intial vocabulary of the model was not suited for Karakalpak, as there were some non-ASCII characters in the writing system (see Table \ref{tab:non-ascii}). We also provide an evaluation of a model without training a separate tokenizer and compare the model's performance with and without additional trained tokens.

\begin{table}[h]
\centering
\begin{tabular}{@{}cccccc@{}}
Á á & Ǵ ǵ & Í ı & Ń ń & Ó ó & Ú ú \
\end{tabular}

\caption{Non-ASCII letters from Karakalpak Latin alphabet. }
\label{tab:non-ascii}

\end{table}

\subsubsection{Vocabulary Expansion}
Following tokenizer training, we augmented the model's vocabulary. This expansion resulted in a total of 269,399 tokens, representing an increase of 13,195 tokens from the original model configuration. We then resized the model's token embeddings and initialized the new embeddings by averaging the embeddings of their constituent subtokens from the original vocabulary.

\begin{table*}[t]
\centering
\small
\begin{tabular}{l|cccccc}
\hline
\textbf{Model} & \textbf{en-kaa} & \textbf{ru-kaa} & \textbf{uz-kaa} & \textbf{kaa-en} & \textbf{kaa-ru} & \textbf{kaa-uz} \\
\hline
madlad-400 & 2.68 / 22.48 & 2.01 / 19.93 & 1.31 / 16.81 & 28.42 / 53.06 & 16.95 / 41.12 & 10.26 / 38.75 \\ 
apertium-uzb-kaa & - & - & 12.26 / 42.27 & - & - & 5.61 / 35.82 \\
google-from-kaz & - & - & - & 20.95 / 44.63 & 13.55 / 36.91 & - \\
google-from-uzb & - & - & - & 21.40 / 45.50 & 13.78 / 37.64 & - \\
nllb-200-600M-from-kaz & - & - & - & 4.32 / 23.35 & 3.12 / 16.86 & 3.91 / 25.26 \\
nllb-200-600M-from-uzb & - & - & - & 8.89 / 32.26 & 5.82 / 26.33 & 4.83 / 29.68 \\
from-to.uz & - & - & \textbf{20.18 / 53.22} & - & - & 11.18 / 41.37 \\
claude-3.5-sonnet & 11.17 / 33.37 & 9.02 / 34.02 & 12.74 / 35.17 & \textbf{37.06 / 61.41} & \textbf{25.70 / 51.23} & \textbf{22.38 / 54.71} \\
\hline
dilmash-raw & 14.37 / 45.65 & 11.41 / 42.99 & 16.16 / 48.88 & 30.01 / 54.81 & 16.34 / 42.01 & 19.19 / 51.92 \\
dilmash & 12.31 / 42.22 & 10.72 / 40.29 & 16.13 / 48.42 & 28.75 / 53.70 & 15.69 / 41.58 & 18.52 / 51.03 \\
dilmash-TIL & \textbf{15.02 / 45.43} & \textbf{12.00 / 42.07} & 17.59 / 49.90 & 32.07 / 56.45 & 17.53 / 43.52 & 19.83 / 52.58 \\
\hline
\end{tabular}
\caption{Evaluation of several models on sacreBLEU/chrF++ across various language pairs with Karakalpak on FLORES+ devtest set. }
\label{tab:translation-performance}
\end{table*}

\subsubsection{Model Variants}
We developed three distinct model variants to evaluate the impact of additional tokens and training data composition:

\begin{enumerate}
    \item \textbf{dilmash-raw\footnote{dilmash \texttt{[dil-mash]} \textit{n. (from Karakalpak)} an oral interpreter}}: This model was trained exclusively on a our own parallel corpus comprising 300,000 sentence pairs in Uzbek, Russian, and English on the original nllb-200-600M.

    \item \textbf{dilmash}: Same as \textbf{dilmash-raw}, but fine-tuned on a model with additional tokens which were trained on a bigger Karakalpak monocorpus.
    
    \item \textbf{dilmash-TIL}: This variant was trained on the same dataset and tokenizer configuration as the \textbf{dilmash}, but supplemented with a strategically sampled subset from the TIL corpus. The sampling strategy was as follows:
    \begin{itemize}
        \item 20\% of parallel datasets containing Uzbek or Kazakh
        \item 5\% of all other datasets in the TIL corpus
    \end{itemize}
\end{enumerate}

To maintain balance with the Karakalpak dataset, we imposed an upper limit of 300,000 sentence pairs on the TIL corpus sample for the \textbf{dilmash-TIL}. This constraint ensured that the Karakalpak data was not overwhelmed by the additional multilingual data, while still allowing for potential improvements in cross-lingual transfer and overall model performance.

With a batch size of 1024 and using the AdaFactor \cite{shazeer2018adafactor} optimizer, we trained each model variant for 3 epochs. We employed a learning rate of 1e-4 with a linear warmup over the first 10\% steps, followed by a constant learning rate schedule. Weight decay was set to 0.01 to help prevent overfitting.

To maximize computational efficiency and enable training on larger batch sizes, we utilized DeepSpeed ZeRO Stage 3 \cite{rasley2020deepspeed} for model parallelism across 16 GPUs. This configuration allowed us to effectively distribute the model parameters and optimize memory usage, facilitating faster training times.



\subsection{Evaluation Metrics}

To evaluate the performance of our translation models, we employ two widely used metrics in machine translation:

\begin{itemize}
\item sacreBLEU \cite{post-2018-call}
\item chrF++ \cite{popovic2017chrf++}
\end{itemize}

sacreBLEU, a standardized BLEU implementation, calculates the similarity between the machine-generated translation and one or more reference translations based on n-gram precision. It addresses inconsistencies in tokenization and BLEU computation across different implementations. chrF++, an extension of the character n-gram F-score, computes the F-score of character n-grams and word unigrams, which is particularly useful for morphologically rich languages, like Karakalpak or Uzbek.

\section{Results and Discussion}

Our evaluation on the FLORES+ Karakalpak devtest reveals several interesting insights into the performance of various translation models. The results, presented in Table \ref{tab:translation-performance}, demonstrate the effectiveness of our proposed models, dilmash, dilmash-raw, and dilmash-TIL, in comparison to existing approaches.

Notably, the dilmash-raw model, which was trained on the original nllb-200-600M without additional tokens, outperforms the dilmash model with expanded vocabulary in most language pairs. This result suggests that the initialization of new token embeddings may have introduced some challenges. Our hypothesis is that the new token embeddings weren't initialized optimally, and before the model could learn good values for them, they may have affected other model parameters. The limited amount of Karakalpak data alone might not have been sufficient for the model to fully compensate for this initial distortion.

The dilmash-TIL model, which incorporates additional multilingual data from the TIL corpus, consistently outperforms both dilmash and dilmash-raw across all language pairs. This improvement is particularly notable in the \textbf{*-kaa} directions, with gains of up to 2.71 BLEU points (en-kaa) compared to dilmash. These results underscore two important points: first, the potential of using related Turkic language data to enhance translation quality for low-resource languages like Karakalpak; and second, that the additional training data and epochs may have allowed the model to better utilize the expanded vocabulary, overcoming the initial challenges faced by the dilmash model. To provide a more qualitative assessment of our models' performance, we have included translation examples in Appendix \ref{sec:translation-examples}.

While expanding the vocabulary can potentially improve model performance, careful consideration must be given to the initialization of new embeddings and the amount of training data available. The success of the dilmash-TIL model suggests that incorporating data from related languages and allowing for longer training periods can help overcome these challenges, ultimately leading to improved translation quality.

Interestingly, the Claude-3.5-sonnet model demonstrates superior performance in the \textbf{kaa-*} directions, surpassing our models by a significant margin. This suggests that large language models may have a particular advantage in understanding content in low-resource languages, possibly due to their extensive pretraining on diverse multilingual data. 

The performance of other baseline models provides additional context. Google Translate when treating Karakalpak as Kazakh or Uzbek, achieves respectable results but falls short of our models and Claude-3.5-sonnet. The NLLB-200-600M model, despite not being originally trained on Karakalpak, shows some ability to transfer knowledge when treating Karakalpak as Uzbek rather than Kazakh. This aligns with linguistic expectations, given the closer relationship between Karakalpak and Uzbek (both in linguistic similarity and writing scripts).

\section{Conclusion}

Our key contributions in this work include:

\begin{enumerate}
    \item Creation of a FLORES+ devtest dataset for Karakalpak.
    \item Development of parallel corpora for Uzbek-Karakalpak, Russian-Karakalpak, and English-Karakalpak, each containing 100,000 sentence pairs.
    \item Open-sourcing of fine-tuned neural models for translation across Uzbek, Russian, English, and Karakalpak languages.
    \item Open-sourcing of scripts for Latin-Cyrillic transliteration for Karakalpak.
\end{enumerate}

Looking ahead, we plan to explore data augmentation techniques to further enhance our models' performance. One promising approach is to leverage the capabilities of Claude-3.5-sonnet for back-translation, potentially expanding our training data with high-quality synthetic examples.

Additionally, we aim to expand our dataset by mining more data from a wider range of books and sources. This will not only increase the volume of our training data but also improve its diversity, potentially leading to more robust and versatile translation models.

\section{Limitations}

While our study presents some advancements in Karakalpak machine translation, several limitations should be noted. First, the relatively small size of our dataset, despite being substantial for a low-resource language, may limit the model's ability to generalize across diverse domains and linguistic contexts. Second, the reliance on machine translation for creating the English-Karakalpak dataset introduces potential biases and errors that could affect model performance. Additionally, our evaluation is primarily based on automatic metrics, which may not fully capture the nuances of translation quality, particularly for a morphologically rich language like Karakalpak. Future work should address these limitations through expanded data collection, human evaluation, and more diverse testing scenarios.

\section{Acknowledgements}
We thank David Dalé for his valuable insights and guidance throughout the project. Our heartfelt appreciation goes to Najimova Perizad and Nurlan Pirjanov for their expertise and assistance with the Karakalpak language. We are also grateful to Atabek Murtazaev and Nurniyazov Ajiniyaz for their support and contributions to this work. Their collective efforts and knowledge have been instrumental in improving the quality of this work.

\clearpage

\bibliography{anthology,custom}
\bibliographystyle{acl_natbib}

\clearpage

\onecolumn

\appendix

\section{Prompt for translating from Russian to English using Claude-3.5-sonnet}
\label{sec:claude-prompt}

\begin{verbatim}

You are a professional translator specializing in Russian to English translations. 
Your task is to translate the given Russian text into English with the highest level 
of accuracy, preserving the original meaning and context. Use proper grammar, 
punctuation, and idiomatic expressions appropriate for English speakers. 
Do not include any additional explanations or commentary; provide only the translated text.

Russian: {sent}
English:

\end{verbatim}

\clearpage

\section{Translation examples from dilmash-TIL}
\label{sec:translation-examples}

\begin{table}[h]
\centering
\begin{tabular}{|p{0.3\textwidth}|p{0.3\textwidth}|p{0.3\textwidth}|}
\hline
\textbf{eng} & \textbf{kaa} & \textbf{dilmash-til (eng → kaa)} \\

\hline

The winter can be deceptively chilly: temperatures rarely go below freezing, but the wind and humidity combine to make it feel colder than what the thermometer says. & Qıs aldamshı dárejede suwıq bolıwı múmkin: temperatura kemnen-kem jaǵdaylarda muzlawdan tómenge túsedi, bıraq samal hám ıǵallıq termometr aytqanınan kóre suwıqlaw seziliw ushın birlesedi. & Qıs júdá suwıq bolıwı múmkin: temperaturanıń buzılıw dárejesinen az ushırasıwı múmkin, biraq samal hám ıallıqtıń birigiwinen termometrdiń aytqanına qaraganda suwiq boladı. \\

\hline 

\textbf{kaa} & \textbf{eng} & \textbf{dilmash-til (kaa → eng)} \\

\hline

Transport joq ekenligi qáweteri oyındı topar tárepdarlarısız jabıq esikler artında ótkeriwge májbúr bolıw múmkinshiligın asırdı. & Fears of lack of transportation raised the possibility that the game would be forced to play behind closed doors without the team's supporters. & Concern over the lack of transportation increased the possibility of being forced to hold the game behind closed doors without group supporters. \\

\hline

\textbf{rus} & \textbf{kaa} & \textbf{dilmash-til (rus → kaa)} \\

\hline

\foreignlanguage{russian}{Также полезно носить кольцо (только не слишком дорогое на вид).} & Júzik taǵıw da paydalı bolıp tabıladı (júdá qımbat kórinetuǵın júzik emes). & Sonday-aq, dóńgelek kiyiwde paydalı boladı (tek júdá qımbat kórinetuın emes). \\

\hline

\textbf{kaa} & \textbf{rus} & \textbf{dilmash-til (kaa → rus)} \\

\hline

Ullı piramida Firaun Xufudı ullılaw ushın jaratılǵan hám kóplegen kishi piramidalar, qábirler hám ibadatcxanalar Xufunıń hayalları hám shańaraq aǵzaların ullılaw ushın qurılǵan. & \foreignlanguage{russian}{Великая Пирамида была создана в честь фараона Хеопса, а много меньших пирамид, гробниц и храмов были построены в честь жён Хеопаса и его родственников.} & \foreignlanguage{russian}{Большая пирамида была создана для восхваления фараона Хуфа, а многие небольшие пирамиды, могилы и храмы были построены для восхваления жен Хуфа и членов его семьи.} \\

\hline

\textbf{uzb} & \textbf{kaa} & \textbf{dilmash-til (uzb → kaa)} \\

\hline

Ayrim atomlar turg‘un bo‘lmagan yadroga ega, bu esa ularning kichkina turtki bilan yoki turtkisiz parchalanishga moyilligini anglatadi. & Ayırım atomlar turaqsız yadrolarǵa iye, yaǵnıy olar azǵantay yamasa hesh qanday túrtki bolmaǵan halda bóleklenedi. & Ayırım atomlar turaqsız yadroga iye, bul bolsa olardıń kishi túrtki menen yaki túrtkisiz bóliniwine beyim ekenligi anlatadı. \\

\hline

\textbf{kaa} & \textbf{uzb} & \textbf{dilmash-til (kaa → uzb)} \\

\hline

Keshesi 150 den 200 ge shekem nusqalar tayarlandı, házirde "Dunlap broadsides" dep ataladı. & Tun davomida 150 dan 200 ga qadar nusxalar tayyorlandi, ular hozirda "Danlep yon zambaraklari" deb ataladi. & Kechasi 150 dan 200 gacha nusxalar tayyorlandi, hozirda "Dunlap broadsides" deb ataladi. \\ 

\hline

\end{tabular}
\caption{Some translation examples of dilmash-TIL model on FLORES+ sentences.}
\label{tab:your-label}
\end{table}

\end{document}